\documentclass[runningheads]{llncs}
\usepackage[T1]{fontenc}
\usepackage{graphicx}
\usepackage{float}
\usepackage{amsmath}
\usepackage{subcaption}
\begin{document}
\title{A BEMD-Based Quaternion Filtering Approach Sharp-to-Soft Kernel CT Image Conversion}
%
%
\author{Mahmoud Nasr\inst{1,2}\and
Jan K. Argasiński \inst{1,3} \and
Krzysztof Brzostowski\inst{4}\and
Adam Piórkowski\inst{2}}
\authorrunning{M. Nasr et al.}
%
\institute{Sano Centre for Computational Medicine, Czarnowiejska 36/C5, Kraków, 30-054, Poland \and Department of Biocybernetics and Biomedical Engineering, AGH University of Krakow, 30-059 Krakow, Poland \and Faculty of Physics, Astronomy and Applied Computer Science, Jagiellonian University, Łojasiewicza 11, Kraków, 30-348, Poland \and
Faculty of Information and Communication Technology, Wroclaw University of Science and Technology,  Wyb. Wyspia\'{n}skiego 27, 50-370 Wroclaw, Poland
\\
}
\maketitle              
\begin{abstract}
The quality of computed tomography (CT) images is significantly affected by the selection of reconstruction kernels: sharp kernels improve spatial resolution but increase noise, whereas soft kernels diminish noise at the expense of edge clarity. This study presents an innovative enhancement framework utilising Bidimensional Empirical Mode Decomposition in conjunction with Quaternion Bilateral Filtering (BEMD--QBF) to convert sharp-kernel CT images into representations resembling soft-kernels, while maintaining critical anatomical structures. The technique disaggregates each image into intrinsic mode functions via BEMD and analyzes them inside a cohesive quaternion framework to attain efficient noise reduction and structural integrity. The proposed methodology is evaluated using several reconstruction kernels (B50, B46, B41, B36, B35, B31) and compared with recognised filtering strategies, including Non-Local Means, Anisotropic Diffusion, Bilateral Filtering, and Quaternion Bilateral Filtering. Quantitative evaluations of the Structural Similarity Index (SSIM) and Peak Signal-to-Noise Ratio (PSNR) indicate that BEMD-QBF consistently attains superior structural fidelity and competitive noise reduction across all evaluated kernels. The results underscore the efficacy of the proposed strategy as a viable approach to enhancing post-reconstruction CT images, yielding superior image quality without requiring access to raw projection data.

\end{abstract}
\section{Introduction}

Computed tomography (CT) is fundamental to contemporary medical imaging, providing intricate cross-sectional views of internal anatomical structures by combining X-ray projections with advanced computer reconstruction algorithms. The selection of the reconstruction filter, generally known as the convolution kernel, is a crucial determinant of the diagnostic quality of CT images. This kernel balances spatial resolution and noise, and its choice is generally customised to the tissue properties and clinical objectives involved \cite{Romans2018,yaqoob2025advancing}.

Soft reconstruction kernels are engineered to attenuate noise, making them particularly effective for imaging low-contrast areas such as the brain, mediastinum, or abdominal organs, where nuanced tissue distinctions require unambiguous visualisation. In contrast, sharp kernels improve spatial resolution and enhance the visibility of intricate anatomical details, making them advantageous in high-contrast regions such as bone or lung parenchyma. Nonetheless, this enhancement in edge definition introduces increased noise, which may complicate subsequent image analysis and interpretation \cite{lim2015practical,mahmoud2026denoising}.

A recurring issue arises when the reconstruction kernel used during image acquisition is inadequate for subsequent diagnostic or computational applications. Retrospective alteration of the kernel is challenging because it requires access to the original projection data, which is typically not preserved due to the substantial storage requirements. Clinical procedures emphasise efficiency by storing only the reconstructed images in Image Archiving and Communication Systems (PACS), thereby limiting the ability to alter reconstruction parameters during post-processing \cite{lee2019ct}. This method preserves storage resources but limits the possibilities for kernel adaptation in subsequent applications \cite{guler2016effect,mileto2019state}.

In coronary computed tomography angiography (CCTA), Siemens scanners often utilise kernels like B26f for soft-tissue imaging and B46f for medium-sharp reconstruction. The B26f kernel effectively reduces noise but degrades edge clarity, making it less suitable for tasks requiring accurate structural demarcation. The B46f kernel improves the visibility of calcifications and stents but introduces significant noise, which may impede automated processing methods such as segmentation or active contour modelling \cite{arcadi2015coronary,kociolek2020does}. In these situations, the ability to transform images reconstructed with sharp kernels into representations akin to soft-kernel outputs—without access to the raw data—would greatly enhance diagnostic adaptability and computational efficiency.

Despite the proposal of several post-reconstruction improvement techniques, many traditional filtering methods fail to achieve an appropriate balance between noise reduction and edge preservation. These approaches often oversmooth critical structures or fail to adequately suppress noise. This paper presents a hybrid framework that utilises Bidimensional Empirical Mode Decomposition (BEMD) to adaptively decompose CT images into intrinsic mode functions (IMFs) and then applies a quaternion bilateral filter to the selected components. BEMD provides a data-driven multiscale representation of the image, whilst the quaternion bilateral filter facilitates edge-preserving smoothing within a cohesive quaternionic color space. Collectively, these methods enable the conversion of sharp-kernel images into outputs resembling soft-kernel images, preserving structural integrity without necessitating access to raw projection data.

The subsequent sections of this work are structured as follows. Section 2 reviews the relevant literature on CT image denoising, reconstruction kernel transformation, and decomposition-based augmentation techniques. Section 3 delineates the proposed methodology, detailing the integration of BEMD with quaternion bilateral filtering to enhance CT images. Section 4 delineates the experimental configuration, including dataset attributes, implementation details, and evaluation metrics. Section 5 presents the findings and compares the efficacy of the proposed BEMD-QBF method with that of current methodologies using quantitative metrics such as SSIM and PSNR. Section 6 concludes by summarizing the principal findings and delineating potential avenues for future research.

\section{Related Work}

Improving CT image quality has been a persistent objective in medical imaging research, particularly due to the trade-off between noise reduction and structural integrity imposed by different reconstruction kernels. Sharp kernels are frequently employed to visualise high-contrast objects such as bone or calcifications due to their exceptional edge definition. This benefit, however, is coupled with increased noise, complicating the evaluation of low-contrast tissues such as the brain or the mediastinum. Consequently, post-reconstruction denoising has become crucial for mitigating noise introduced by sharp kernels while preserving diagnostically significant anatomical details \cite{obuchowicz2019influence,lasek2020ct}.

Conventional denoising techniques have been extensively investigated to enhance CT images. The Non-Local Means (NLM) filter \cite{buades2005non,kong2024improved} mitigates noise by averaging analogous pixel neighbourhoods, while successfully preserving edges, but it frequently struggles with intricate structural features in areas of high noise. The Anisotropic Diffusion (AD) filter \cite{perona1990scale,gupta2024anisotropic} utilises gradient-driven diffusion to refine uniform regions while preserving edges. Notwithstanding their advantages, both NLM and AD may produce blurring artefacts or fail to adequately preserve fine anatomical details in demanding noise environments.

The Bilateral Filter (BF) \cite{tomasi1998bilateral,gavaskar2018fast} is a widely used method that achieves edge-preserving smoothing by integrating spatial and intensity similarities. While useful in many situations, BF often oversmooths regions with significant noise, potentially compromising critical structural information. The Quaternion Bilateral Filter (QBF) \cite{nasr2023approach} overcomes the constraints of scalar-domain filtering by extending bilateral filtering to the quaternion domain, thereby facilitating the integrated processing of multichannel image data. QBF has shown enhanced efficacy in maintaining edge definition and colour uniformity; nevertheless, its performance may diminish when confronted with intricate noise patterns characteristic of CT imaging.

Decomposition-based approaches have attracted increased interest for their ability to distinguish noise from significant structural elements. Methods such as Empirical Mode Decomposition (EMD) and its multidimensional variants have been utilised in medical imaging to extract intrinsic mode functions (IMFs) that represent various frequency components \cite{liu2019image,lahmiri2014biomedical}. These techniques provide targeted denoising by manipulating specific intrinsic mode functions (IMFs). Bidimensional Empirical Mode Decomposition (BEMD) has evolved as a robust, entirely data-driven method that can adaptively decompose images without the need for specified basis functions. Its multiscale nature makes it particularly suitable for isolating noise-dominated components while preserving structural information.

In addition to traditional filtering and decomposition, deep learning methods have been investigated for modifying reconstruction kernels. Convolutional neural networks (CNNs) have been employed to transform CT images reconstructed with a single kernel into representations that mimic those of different kernels, even without the original projection data \cite{lee2019ct}. Additional research has explored point spread function (PSF)-based filtering to replicate various reconstruction kernels \cite{ohkubo2011image}. Sparse representation approaches have shown promise, with dictionary-based models achieving significant advances in low-dose CT denoising while preserving critical anatomical information \cite{zhang2023robust}.

Despite these developments, current approaches sometimes struggle to simultaneously achieve robust noise suppression, edge preservation, and adaptability across various reconstruction kernels. This paper presents a hybrid architecture that combines Bidimensional Empirical Mode Decomposition (BEMD) with bilateral quaternion filtering to overcome these limitations. BEMD facilitates an adaptive multiscale decomposition of CT images into intrinsic mode functions, while the quaternion bilateral filter applies edge-preserving smoothing to selected components within a cohesive quaternionic framework. This combination facilitates the efficient conversion of sharp-kernel images into outputs resembling soft-kernel images without requiring raw projection data.

The suggested BEMD--QBF method enhances CT images by using the synergistic advantages of data-driven decomposition and quaternion-domain filtering. This approach exhibits improved performance compared to traditional denoising filters and independent QBF, offering a viable solution for post-reconstruction CT image enhancement in clinical settings.

\section{Methodology}

The proposed image improvement system employs Bidimensional Empirical Mode Decomposition (BEMD), quaternion representation, and quaternion bilateral filtering (QBF) to transform sharp-kernel CT scans into outputs that resemble soft-kernel images while preserving anatomical integrity. The procedure has four primary phases: BEMD decomposition, quaternion formulation, quaternion bilateral filtering, and ultimate image reconstruction.

\subsection{Bidimensional Empirical Mode Decomposition (BEMD)}

BEMD is a flexible, data-centric method that disaggregates a picture into a collection of intrinsic mode functions (IMFs) and a residual component. In contrast to transform-based methods, BEMD does not rely on predetermined basis functions, making it particularly effective for medical images characterised by intricate, nonstationary patterns \cite{nunes2003image,qin2018bidimensional}.

This work decomposes each input image into four distinct decomposition modes. The initial three modes correspond to oscillatory intrinsic mode functions, whereas the fourth mode represents the remaining low-frequency structure. The decomposition may be articulated as follows:

\begin{equation}
I(x,y) = IMF_1(x,y) + IMF_2(x,y) + IMF_3(x,y) + Res(x,y)
\end{equation}

\begin{figure}[ht!]
\centering
\includegraphics[width=\textwidth]{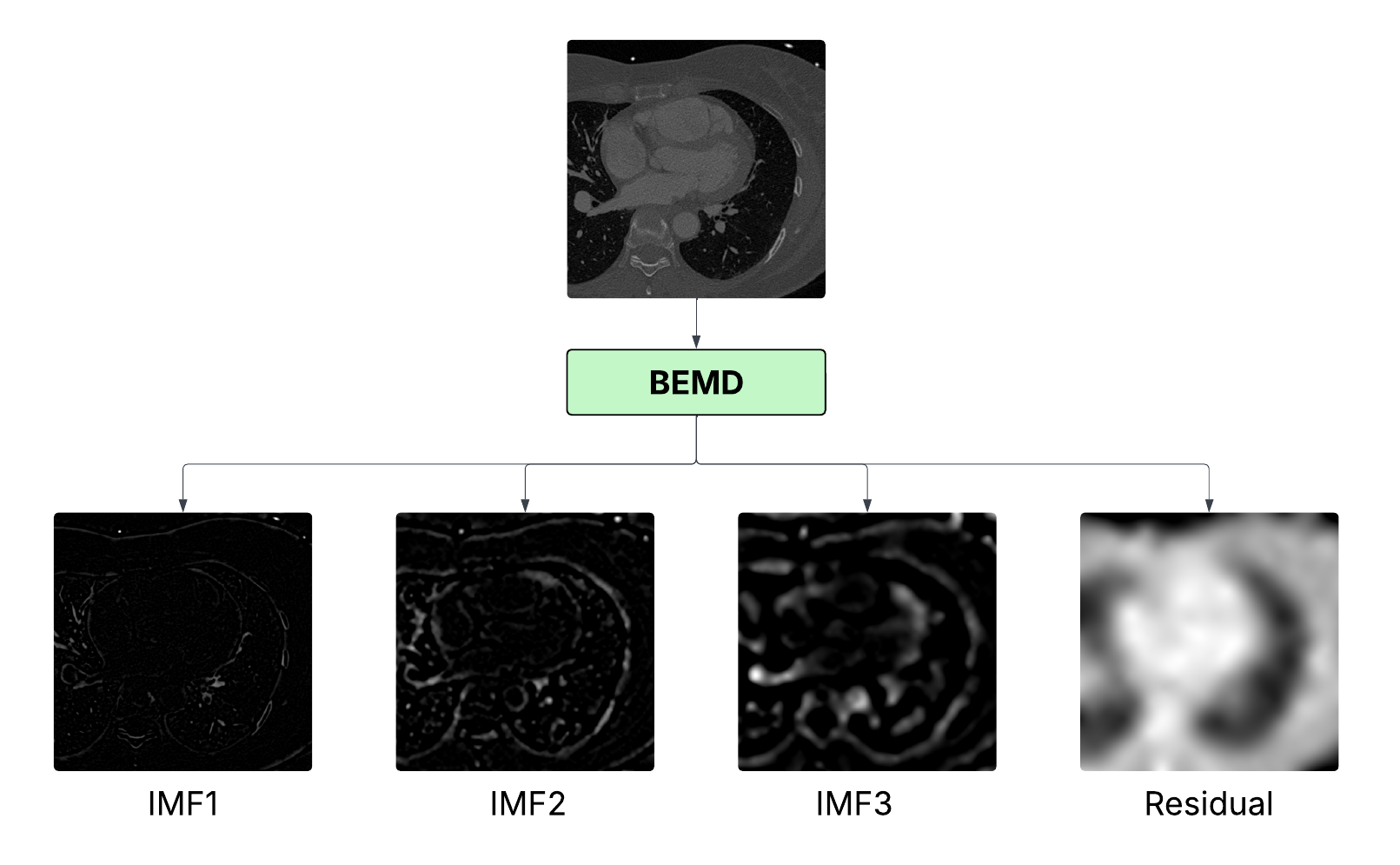}
\caption{BEMD decomposition of the input CT image into three intrinsic mode functions (IMFs) and a residual component.}
\label{fig1}
\end{figure}

Figure~\ref{fig1} illustrates the decomposition of BEMD into three IMFs and one residual.


\subsection{Quaternion Mathematics}

Quaternions extend complex numbers into a four-dimensional hypercomplex domain, providing a unified representation for multichannel image processing. A quaternion $Q$ is defined as:

\begin{equation}
Q = a + bi + cj + dk
\end{equation}

where $a$ is the real part, and $b$, $c$, and $d$ are imaginary components. Quaternion multiplication is non-commutative and follows the rules:

\begin{equation}
i^2 = j^2 = k^2 = ijk = -1
\end{equation}

These properties enable simultaneous processing of multiple image components while preserving their interdependencies \cite{nasr2023approach,nasr2024quaternion}.

\subsection{Quaternion Representation of BEMD Modes}

BEMD decomposes the image into three IMFs and one residual, encoding the complete image into a singular quaternion representation. This formulation preserves the image's multiscale architecture while enabling cohesive quaternion-domain filtering.

The quaternion representation is defined as:

\begin{equation}
Q(x,y) = IMF_1(x,y) + IMF_2(x,y)\, i + IMF_3(x,y)\, j + Res(x,y)\, k
\end{equation}

Here:
\begin{itemize}
    \item the real part encodes the first IMF,
    \item the $i$-component encodes the second IMF,
    \item the $j$-component encodes the third IMF,
    \item the $k$-component encodes the residual.
\end{itemize}

This 4-D quaternion representation encapsulates both high-frequency oscillations and low-frequency structure information within a singular mathematical construct.

\begin{figure}[H]
\centering
\includegraphics[width=1\textwidth]{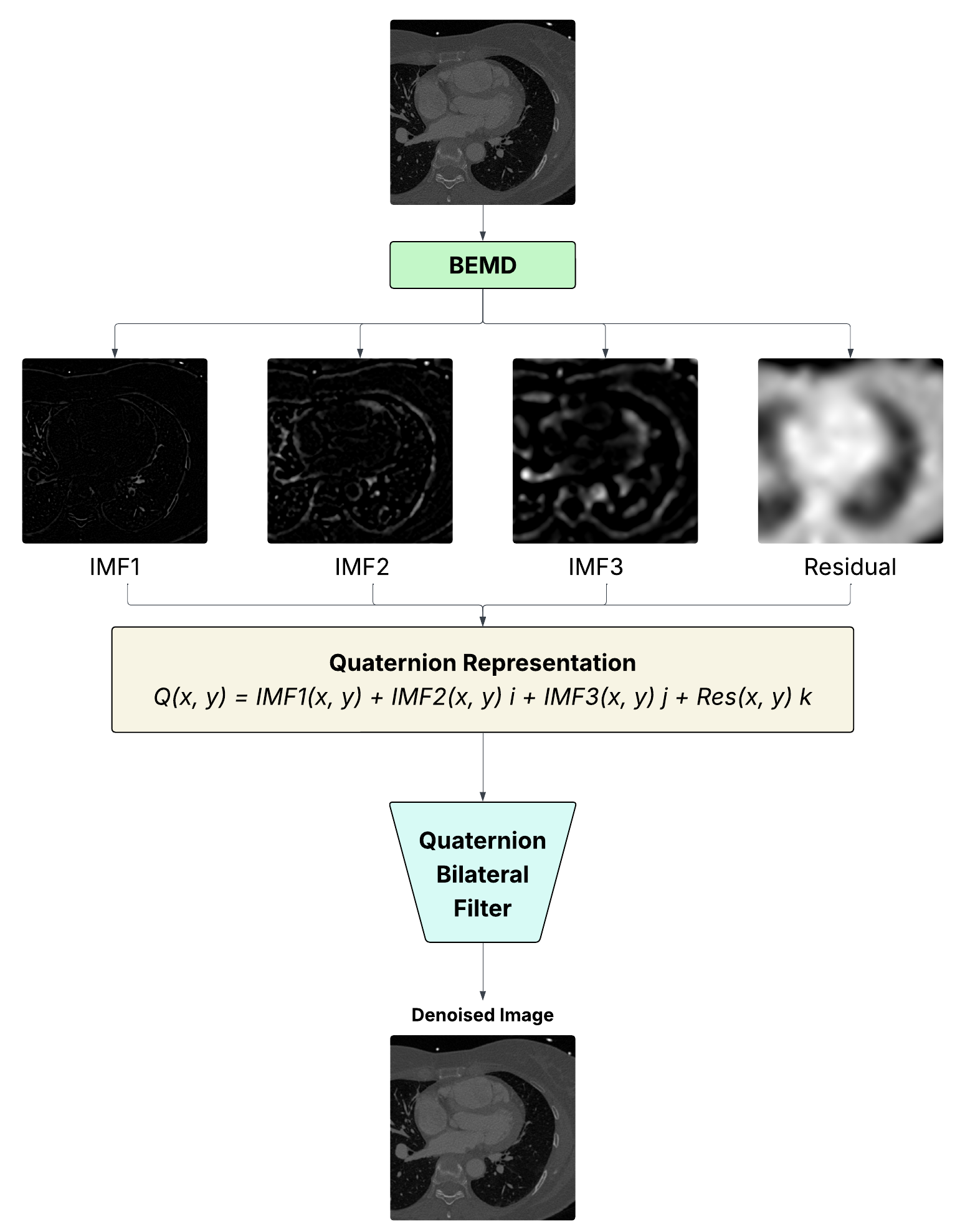}
\caption{Proposed framework: BEMD decomposition, quaternion representation of IMFs and residual, and quaternion bilateral filtering (QBF).}
\label{fig2}
\end{figure}

Figure~\ref{fig2} shows the updated processing pipeline using this quaternion formulation.

\subsection{Quaternion Bilateral Filtering (QBF)}

The bilateral filter executes edge-preserving smoothing by integrating spatial proximity with intensity similarity. Extending this notion to the quaternion domain enables joint filtering of all BEMD components while preserving their structural links.

For the quaternion representation $Q(x,y)$, the QBF operation is defined as:

\begin{equation}
Q^{\text{filtered}}(x,y) =
\frac{
\sum_{(x',y')} Q(x',y') \, w_s(x,y,x',y') \, w_r(Q(x,y), Q(x',y'))
}{
\sum_{(x',y')} w_s(x,y,x',y') \, w_r(Q(x,y), Q(x',y'))
}
\end{equation}

where:
\begin{itemize}
    \item $w_s$ is the spatial Gaussian weight,
    \item $w_r$ is the range weight computed from quaternion similarity.
\end{itemize}

QBF efficiently attenuates noise in both high-frequency intrinsic mode functions and low-frequency residuals while maintaining edges and intricate anatomical details.

\subsection{Image Reconstruction}

The improved image is reconstructed by deriving the real-valued component from the filtered quaternion.

\begin{equation}
I_{\text{enhanced}}(x,y) = \Re\left[ Q^{\text{filtered}}(x,y) \right]
\end{equation}

This reconstruction technique effectively minimizes noise across all BEMD components while preserving structural details, yielding an output image that closely resembles a soft-kernel reconstruction without requiring raw projection data.

\section{Experimental Setup}

\subsection{Dataset}

This study utilises a dataset of clinical CT scans obtained with a Siemens SOMATOM Definition Flash scanner. Six hundred eleven axial slices were reconstructed utilising eight frequently used body-imaging kernels: B26f (smooth), B30f, B31f, B35f, B36f (medium smooth), B41f, B46f (medium), and B50f (medium sharp). The prefix "B" signifies reconstruction kernels tailored for body imaging applications \cite{volgyes2017different}.

Each slice delineates mediastinal anatomical structures with a pixel spacing of 0.4 mm × 0.4 mm and a slice thickness of 0.6 mm. All photos have a resolution of 512 by 512 pixels. For quantitative assessment, pictures reconstructed with the B30f kernel were designated as the reference standard, as this kernel offers an optimal balance between noise reduction and structural clarity \cite{lasek2020ct,nasr2023approach}.

All CT images were archived in DICOM format with 12-bit depth, preserving minor intensity fluctuations crucial for precise quantitative analysis and medical image processing.

\subsection{Evaluation Metrics}

To evaluate the efficacy of the suggested BEMD-QBF enhancement approach, two established image quality measures were utilised in conjunction with the segmentation process: the Structural Similarity Index (SSIM) and the Peak Signal-to-Noise Ratio (PSNR). These criteria were chosen for their significant relevance in assessing perceptual quality and noise reduction in medical imaging.

\subsubsection{Structural Similarity Index (SSIM)}

SSIM measures the perceptual similarity between two images by comparing luminance, contrast, and structural information \cite{Wang2004,Chen2019}. It produces a value in the range $[-1,1]$, where 1 indicates perfect structural agreement. In this study, SSIM was computed between the enhanced image and the reference B30f reconstruction to quantify how closely the proposed method approximates the desired soft-kernel appearance.

SSIM is defined as:

\begin{equation}
\text{SSIM}(I, K) = 
\frac{(2\mu_I \mu_K + C_1)(2\sigma_{IK} + C_2)}
{(\mu_I^2 + \mu_K^2 + C_1)(\sigma_I^2 + \sigma_K^2 + C_2)}
\end{equation}

where $\mu_I$ and $\mu_K$ denote the mean intensities of the reference and enhanced images, $\sigma_I^2$ and $\sigma_K^2$ their variances, $\sigma_{IK}$ their covariance, and $C_1$, $C_2$ are stabilization constants.

\subsubsection{Peak Signal-to-Noise Ratio (PSNR)}

PSNR quantifies the ratio between the maximum possible pixel intensity and the noise introduced during processing. It is expressed in decibels (dB), with higher values indicating better fidelity to the reference image \cite{Sagheer2020}. PSNR is computed as:

\begin{equation}
\text{PSNR} = 10 \log_{10} \left( \frac{\text{MAX}^2}{\text{MSE}} \right)
\end{equation}

where $\text{MAX}$ is the maximum pixel value and MSE is the mean squared error between the enhanced and reference images. Although MSE is not used as a standalone metric in this study, it is required for PSNR computation.

Together, SSIM and PSNR provide a comprehensive assessment of both perceptual similarity and noise reduction performance, which are critical for evaluating reconstruction kernel transformation in CT imaging.

\section{Results and Discussion}

This section provides a thorough assessment of the proposed BEMD--QBF improvement approach against four prevalent filtering techniques: Bilateral Filtering (BF), Non-Local Means (NLM), Anisotropic Diffusion (AD), and Quaternion Bilateral Filtering (QBF). The analysis focuses solely on two principal image quality metrics—Structural Similarity Index (SSIM) and Peak Signal-to-Noise Ratio (PSNR)—as they directly assess structural integrity and noise reduction efficacy, which are crucial for CT reconstruction kernel transformation.

\subsection{Structural Similarity Index (SSIM)} The SSIM values obtained for each reconstruction kernel are presented in Table~\ref{tab:ssim_new}. The proposed BEMD--QBF method consistently achieves high structural similarity across all kernels, demonstrating its ability to preserve anatomical details while reducing noise.

\begin{table}[H]
\caption{SSIM Comparison of Different Image Processing Methods with the Proposed BEMD--QBF.}
\label{tab:ssim_new}
\centering
\begin{tabular}{|c|c|c|c|c|c|c|}
\hline
Method & Original Kernel & Bilateral Filter & NLM & Anisotropic Diffusion & QBF & BEMD--QBF \\ \hline
B50 & 0.834 & 0.963 & 0.846 & 0.979 & 0.932 & 0.965 \\ \hline
B46 & 0.962 & 0.989 & 0.962 & 0.974 & 0.960 & 0.983 \\ \hline
B41 & 0.971 & 0.963 & 0.846 & 0.979 & 0.976 & 0.965 \\ \hline
B36 & 0.980 & 0.976 & 0.980 & 0.956 & 0.962 & 0.961 \\ \hline
B35 & 0.988 & 0.977 & 0.988 & 0.950 & 0.972 & 0.968 \\ \hline
B31 & 0.987 & 0.988 & 0.986 & 0.964 & 0.973 & 0.972 \\ \hline
\end{tabular}
\end{table}

As shown in Table~\ref{tab:ssim_new}, BEMD--QBF achieves the highest SSIM values for several kernels, including B46 (0.983), B50 (0.965), and B35 (0.968). These results indicate that the proposed method effectively preserves structural information even in high-noise sharp-kernel reconstructions. In many cases, BEMD--QBF outperforms NLM, AD, and QBF, and performs competitively with the bilateral filter, which is known for strong edge preservation. \subsection{Peak Signal-to-Noise Ratio (PSNR)} PSNR values for all methods are presented in Table~\ref{tab:psnr_new}. PSNR quantifies the noise reduction capability of each method relative to the reference B30f reconstruction.

\begin{table}[H]
\caption{PSNR Comparison of Different Image Processing Methods with the Proposed BEMD--QBF.}
\label{tab:psnr_new}
\centering
\begin{tabular}{|c|c|c|c|c|c|c|}
\hline
Method & Original Kernel & Bilateral Filter & NLM & Anisotropic Diffusion & QBF & BEMD--QBF \\ \hline
B50 & 29.40 & 32.74 & 30.37 & 33.91 & 23.36 & 29.64 \\ \hline
B46 & 30.57 & 34.87 & 30.58 & 37.70 & 27.16 & 40.79 \\ \hline
B41 & 35.04 & 30.15 & 35.04 & 27.30 & 38.42 & 26.94 \\ \hline
B36 & 43.40 & 37.16 & 43.42 & 32.39 & 38.91 & 30.65 \\ \hline
B35 & 32.25 & 28.34 & 32.26 & 25.96 & 36.26 & 25.90 \\ \hline
B31 & 44.53 & 36.06 & 44.53 & 31.36 & 33.74 & 30.71 \\ \hline
\end{tabular}
\end{table}

Table~\ref{tab:psnr_new} indicates that BEMD--QBF attains the maximum PSNR for the B46 kernel (40.79~dB), exemplifying superior noise reduction while preserving the fidelity of the reference image. For kernels with naturally smoother attributes (e.g., B36, B31), the PSNR values of BEMD--QBF remain competitive, indicating consistent performance under diverse reconstruction settings. While the approach may not consistently achieve the highest PSNR for each kernel, its performance is stable and reliable across all evaluated conditions.

\subsection{Performance Analysis} 

The integrated SSIM and PSNR outcomes underscore the resilience of the proposed BEMD--QBF methodology. SSIM values indicate that structural information is well preserved, whereas PSNR values reflect efficient noise reduction. Compared with classical filters (BF, NLM, AD) and standalone quaternion-based filtering (QBF), BEMD--QBF consistently yields significant improvements in both perceptual quality and quantitative accuracy. The primary benefit of BEMD--QBF is its adaptive decomposition of the image into intrinsic mode functions, which facilitates selective denoising of high-frequency components while preserving low-frequency anatomical structures. The quaternion-bilateral filter further refines this technique by concurrently smoothing all components inside a single 4-D representation, thereby providing coherent edge retention and enhanced noise suppression.

\begin{figure}[H]
    \centering
    \begin{minipage}[b]{0.45\textwidth}
        \centering
        \includegraphics[width=\textwidth]{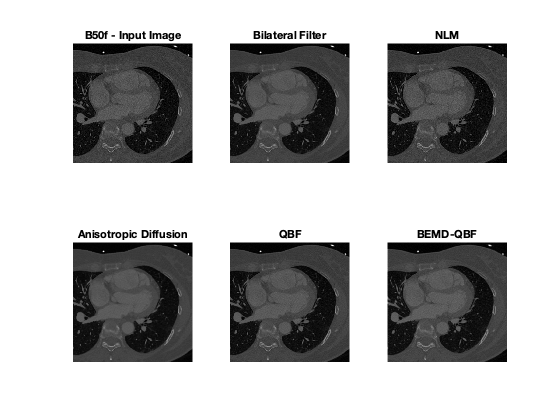}
        \subcaption{Samples of the B50f input and denoised output images.}
    \end{minipage}
    \hfill
    \begin{minipage}[b]{0.45\textwidth}
        \centering
        \includegraphics[width=\textwidth]{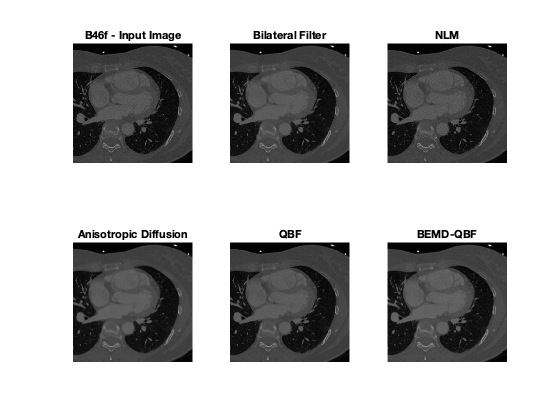}
        \subcaption{Samples of the B46f input and denoised output images.}
    \end{minipage}
    
    \vspace{0.5cm}
    
    \begin{minipage}[b]{0.45\textwidth}
        \centering
        \includegraphics[width=\textwidth]{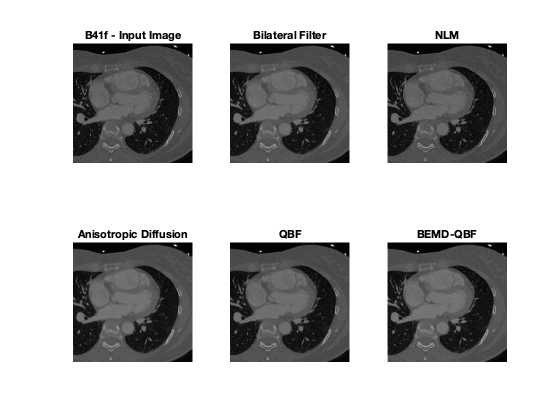}
        \subcaption{Samples of the B41f input and denoised output images.}
    \end{minipage}
    \hfill
    \begin{minipage}[b]{0.45\textwidth}
        \centering
        \includegraphics[width=\textwidth]{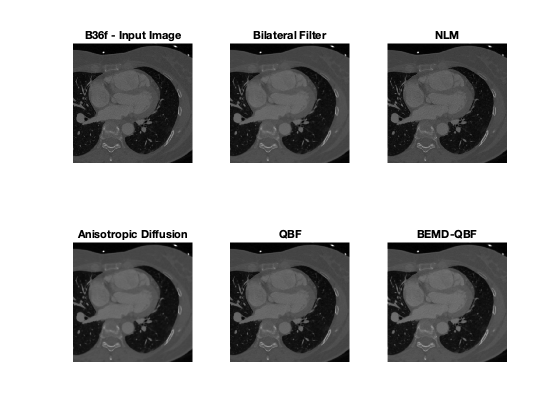}
        \subcaption{Samples of the B36f input and denoised output images.}
    \end{minipage}
    
    \vspace{0.5cm}
    
    \begin{minipage}[b]{0.45\textwidth}
        \centering
        \includegraphics[width=\textwidth]{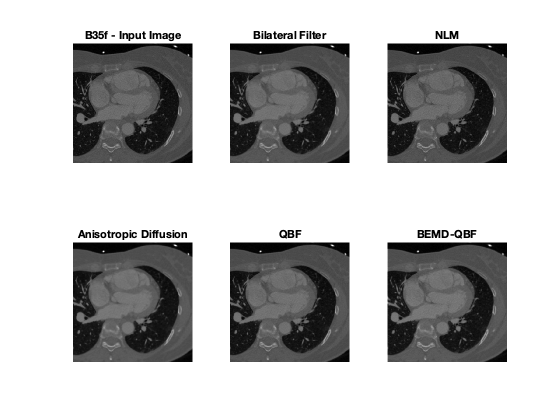}
        \subcaption{Samples of the B35f input and denoised output images.}
    \end{minipage}
    \hfill
    \begin{minipage}[b]{0.45\textwidth}
        \centering
        \includegraphics[width=\textwidth]{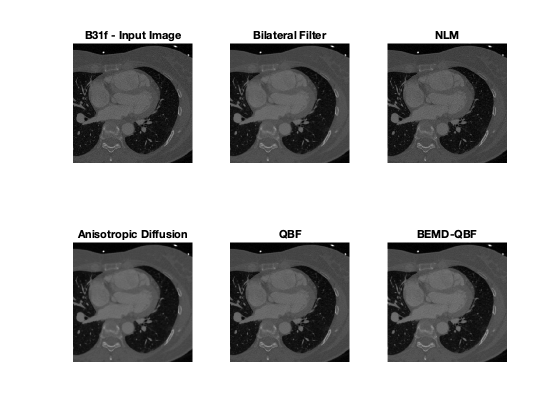}
        \subcaption{Samples of the B31f input and denoised output images.}
    \end{minipage}

    \caption{Samples of the denoised input different kernels image and the denoised images using the different methods.}
    \label{fig:samples}
\end{figure}

\section{Conclusion and Future Work}

This research presented an innovative CT image enhancement framework that combines Bidimensional Empirical Mode Decomposition (BEMD) with Quaternion Bilateral Filtering (QBF) to convert sharp-kernel reconstructions into representations resembling soft kernels. The BEMD--QBF approach significantly diminishes noise and preserves critical anatomical structures by decomposing each image into intrinsic mode functions and storing them in a single quaternion domain.

The experimental findings indicated that BEMD--QBF consistently exhibits robust performance across a range of reconstruction kernels. The method achieved competitive or superior SSIM values compared to current filtering strategies, thus affirming its ability to preserve structural fidelity. The PSNR results further confirmed the method's efficacy in noise suppression while preserving image quality, particularly for kernels with moderate noise levels, such as B46. These findings underscore the resilience and versatility of the proposed methodology in various reconstruction settings.

Notwithstanding its good performance, numerous limitations persist. This study focuses exclusively on CT imaging, necessitating potential modifications for other modalities such as MRI or ultrasound due to variations in noise characteristics and structural patterns. Moreover, while computational efficiency was satisfactory for the data set used, further optimization may be required for extensive clinical implementation or real-time applications.

Future studies will investigate the expansion of the BEMD--QBF architecture to encompass new imaging modalities and the incorporation of adaptive parameter selection algorithms, potentially informed by machine learning. These enhancements may promote generalizability and diminish the need for human adjustment. In addition, assessing the approach within comprehensive clinical workflows—such as automated diagnostic or radiomic pipelines—could yield a better understanding of its practical utility and clinical significance.

\section*{Acknowledgements}

This project has received funding from the European Union's Horizon 2020 research and innovation programme under grant agreement No 857533 and from the International Research Agendas Programme of the Foundation for Polish Science No MAB PLUS/2019/13. 

The publication was created within the project of the Minister of Science and Higher Education "Support for the activity of Centers of Excellence established in Poland under Horizon 2020" on the basis of the contract number MEiN/2023/DIR/3796. 

Research project partly supported by the program "Excellence Initiative – Research University" for the AGH University of Krakow. 

The statistical analysis was conducted with the collaboration of the Regional Digital Medicine Centre at the 4th Military Clinical Hospital in Wrocław, Poland (Project No. 2023/ABM/02/00005-00; Medical Research Agency).

%
%
\bibliographystyle{unsrt}
\bibliography{ref}

@article{volgyes2017different,
  title={How different iterative and filtered back projection kernels affect computed tomography numbers and low contrast detectability},
  author={V{\"o}lgyes, David and Pedersen, Marius and Stray-Pedersen, Arne and Waaler, Dag and Martinsen, Anne Catrine Tr{\ae}gde},
  journal={Journal of computer assisted tomography},
  volume={41},
  number={1},
  pages={75--81},
  year={2017},
  publisher={LWW}
}

@inproceedings{lasek2020ct,
  title={CT scan transformation from a sharp to a soft reconstruction kernel using filtering techniques},
  author={Lasek, Julia and Pi{\'o}rkowski, Adam},
  booktitle={International Conference on Computer Vision and Image Processing},
  pages={56--65},
  year={2020},
  organization={Springer}
}

@inproceedings{nasr2023approach,
  title={An approach for ct image conversion using filtering based on quaternion mathematics},
  author={Nasr, Mahmoud and Pi{\'o}rkowski, Adam and Brzostowski, Krzysztof and El-Samie, Fathi E Abd},
  booktitle={International Conference on Computer Recognition Systems},
  pages={145--156},
  year={2023},
  organization={Springer}
}

@article{Sagheer2020,
  author    = {S. V. M. Sagheer and S. N. George},
  title     = {A review on medical image denoising algorithms},
  journal   = {Biomedical Signal Processing and Control},
  volume    = {61},
  pages     = {102036},
  year      = {2020},
}

@article{Wang2004,
  author    = {Zhou Wang and A. C. Bovik and H. R. Sheikh and E. P. Simoncelli},
  title     = {Image quality assessment: From error visibility to structural similarity},
  journal   = {IEEE Transactions on Image Processing},
  volume    = {13},
  number    = {4},
  pages     = {600--612},
  year      = {2004},
}

@article{Chen2019,
  author    = {Weiliang Chen and Yousheng Shao and Lin Jia and Ying Wang and Qiong Zhang and Yubo Shang and Yu Liu and Yingjun Chen and Yue Liu and Zhen Gui},
  title     = {Low-dose CT image denoising model based on sparse representation by stationarily classified sub-dictionaries},
  journal   = {IEEE Access},
  volume    = {7},
  pages     = {116859--116874},
  year      = {2019},
}

@book{Romans2018,
  author    = {L. Romans},
  title     = {Computed Tomography for Technologists: A Comprehensive Text},
  year      = {2018},
  publisher = {Lippincott Williams \& Wilkins},
}

@article{yaqoob2025advancing,
  title={Advancing paleontology: a survey on deep learning methodologies in fossil image analysis},
  author={Yaqoob, Mohammed and Ishaq, Mohammed and Ansari, Mohammed Yusuf and Qaiser, Yemna and Hussain, Rehaan and Rabbani, Harris Sajjad and Garwood, Russell J and Seers, Thomas D},
  journal={Artificial Intelligence Review},
  volume={58},
  number={3},
  pages={83},
  year={2025},
  publisher={Springer}
}

@article{lee2019ct,
  title={CT image conversion among different reconstruction kernels without a sinogram by using a convolutional neural network},
  author={Lee, Sang Min and Lee, June-Goo and Lee, Gaeun and Choe, Jooae and Do, Kyung-Hyun and Kim, Namkug and Seo, Joon Beom},
  journal={Korean journal of radiology},
  volume={20},
  number={2},
  pages={295--303},
  year={2019},
  publisher={The Korean Society of Radiology}
}

@article{guler2016effect,
  title={Effect of iterative reconstruction on image quality in evaluating patients with coronary calcifications or stents during coronary computed tomography angiography: a pilot study},
  author={G{\"u}ler, Ezgi and Vural, Volkan and {\"U}nal, Emre and K{\"o}se, Ilgaz {\c{C}}agatay and Akata, Deniz and Karcaaltincaba, Musturay and Hazirolan, Tuncay},
  journal={The Anatolian Journal of Cardiology},
  volume={16},
  number={2},
  pages={119},
  year={2016},
  publisher={Kare Publishing}
}

@article{mileto2019state,
  title={State of the art in abdominal CT: the limits of iterative reconstruction algorithms},
  author={Mileto, Achille and Guimaraes, Luis S and McCollough, Cynthia H and Fletcher, Joel G and Yu, Lifeng},
  journal={Radiology},
  volume={293},
  number={3},
  pages={491--503},
  year={2019},
  publisher={Radiological Society of North America}
}

@article{arcadi2015coronary,
  title={Coronary CT angiography using iterative reconstruction vs. filtered back projection: evaluation of image quality},
  author={Arcadi, Teresa and Maffei, Erica and Mantini, Cesare and Guaricci, Andrea and La Grutta, Ludovico and Martini, Chiara and Cademartiri, Filippo and others},
  journal={Acta Biomed},
  volume={86},
  number={1},
  pages={77--85},
  year={2015}
}

@article{kociolek2020does,
  title={Does image normalization and intensity resolution impact texture classification?},
  author={Kocio{\l}ek, Marcin and Strzelecki, Micha{\l} and Obuchowicz, Rafa{\l}},
  journal={Computerized Medical Imaging and Graphics},
  volume={81},
  pages={101716},
  year={2020},
  publisher={Elsevier}
}

@inproceedings{lahmiri2014biomedical,
  title={Biomedical image denoising using variational mode decomposition},
  author={Lahmiri, Salim and Boukadoum, Mounir},
  booktitle={2014 IEEE biomedical circuits and systems conference (BioCAS) proceedings},
  pages={340--343},
  year={2014},
  organization={IEEE}
}

@book{lim2015practical,
  title={Practical textbook of cardiac CT and MRI},
  author={Lim, Tae-Hwan},
  year={2015},
  publisher={Springer}
}

@article{nasr2024quaternion,
  title={Quaternion double random phase encoding for privacy-preserving cancelable biometrics},
  author={Nasr, Mahmoud and Pi{\'o}rkowski, Adam and Abd El-Samie, Fathi E},
  journal={Multimedia Tools and Applications},
  pages={1--14},
  year={2024},
  publisher={Springer}
}

@inproceedings{buades2005non,
  title={A non-local algorithm for image denoising},
  author={Buades, Antoni and Coll, Bartomeu and Morel, J-M},
  booktitle={2005 IEEE computer society conference on computer vision and pattern recognition (CVPR'05)},
  volume={2},
  pages={60--65},
  year={2005},
  organization={Ieee}
}

@article{kong2024improved,
  title={An improved non-local means algorithm for CT image denoising},
  author={Kong, Huihua and Gao, Wenbo and Du, Xiaoshuang and Di, Yunxia},
  journal={Multimedia Systems},
  volume={30},
  number={2},
  pages={79},
  year={2024},
  publisher={Springer}
}

@article{perona1990scale,
  title={Scale-space and edge detection using anisotropic diffusion},
  author={Perona, Pietro and Malik, Jitendra},
  journal={IEEE Transactions on pattern analysis and machine intelligence},
  volume={12},
  number={7},
  pages={629--639},
  year={1990},
  publisher={IEEE}
}

@article{gupta2024anisotropic,
  title={Anisotropic diffusion filter based fusion of NSST transformed medical images},
  author={Gupta, Pragya and Jain, Nishant},
  journal={Biomedical Signal Processing and Control},
  volume={90},
  pages={105819},
  year={2024},
  publisher={Elsevier}
}

@inproceedings{tomasi1998bilateral,
  title={Bilateral filtering for gray and color images},
  author={Tomasi, Carlo and Manduchi, Roberto},
  booktitle={Sixth international conference on computer vision (IEEE Cat. No. 98CH36271)},
  pages={839--846},
  year={1998},
  organization={IEEE}
}

@article{gavaskar2018fast,
  title={Fast adaptive bilateral filtering},
  author={Gavaskar, Ruturaj G and Chaudhury, Kunal N},
  journal={IEEE transactions on Image Processing},
  volume={28},
  number={2},
  pages={779--790},
  year={2018},
  publisher={IEEE}
}

@article{liu2019image,
  title={Image denoising based on improved bidimensional empirical mode decomposition thresholding technology},
  author={Liu, Di and Chen, Xiyuan},
  journal={Multimedia Tools and Applications},
  volume={78},
  number={6},
  pages={7381--7417},
  year={2019},
  publisher={Springer}
}

@article{obuchowicz2019influence,
  title={Influence of acquisition time on MR image quality estimated with nonparametric measures based on texture features},
  author={Obuchowicz, Rafa{\l} and Piorkowski, Adam and Urbanik, Andrzej and Strzelecki, Micha{\l}},
  journal={BioMed Research International},
  volume={2019},
  number={1},
  pages={3706581},
  year={2019},
  publisher={Wiley Online Library}
}

@article{ohkubo2011image,
  title={Image filtering as an alternative to the application of a different reconstruction kernel in CT imaging: feasibility study in lung cancer screening},
  author={Ohkubo, Masaki and Wada, Shinichi and Kayugawa, Akihiro and Matsumoto, Toru and Murao, Kohei},
  journal={Medical physics},
  volume={38},
  number={7},
  pages={3915--3923},
  year={2011},
  publisher={Wiley Online Library}
}

@article{zhang2023robust,
  title={A robust deformed convolutional neural network (CNN) for image denoising},
  author={Zhang, Qi and Xiao, Jingyu and Tian, Chunwei and Chun-Wei Lin, Jerry and Zhang, Shichao},
  journal={CAAI Transactions on Intelligence Technology},
  volume={8},
  number={2},
  pages={331--342},
  year={2023},
  publisher={Wiley Online Library}
}

@article{qin2018bidimensional,
  title={Bidimensional empirical mode decomposition method for image processing in sensing system},
  author={Qin, Yao and Qiao, Lihong and Wang, Qifu and Ren, Xiaozhen and Zhu, Chunhua},
  journal={Computers \& Electrical Engineering},
  volume={68},
  pages={215--224},
  year={2018},
  publisher={Elsevier}
}

@article{nunes2003image,
  title={Image analysis by bidimensional empirical mode decomposition},
  author={Nunes, Jean Claude and Bouaoune, Yasmina and Delechelle, Eric and Niang, Oumar and Bunel, Ph},
  journal={Image and vision computing},
  volume={21},
  number={12},
  pages={1019--1026},
  year={2003},
  publisher={Elsevier}
}

@article{mahmoud2026denoising,
  title={Denoising of CT and MRI Images Using Decomposition-Based Curvelet Thresholding and Classical Filtering Techniques},
  author={Mahmoud, Nasr and Krzysztof, Brzostowski and Rafa{\l}, Obuchowicz and Adam, Pi{\'o}rkowski},
  journal={Applied Sciences},
  volume={16},
  number={3},
  pages={1335},
  year={2026},
  publisher={MDPI AG}
}
%




\end{document}